\documentclass[journal]{IEEEtran}      

\IEEEoverridecommandlockouts           

\usepackage{graphicx}
\usepackage{graphics}
\usepackage{amsmath}
\usepackage{amssymb}
\usepackage{soul}
\usepackage{algorithm}
\usepackage{algpseudocode}
\usepackage{cite}
\usepackage{hyperref}
\usepackage[caption=false]{subfig}
\usepackage{booktabs}
\usepackage{siunitx}
\usepackage{float}
\usepackage{multirow}
\usepackage{bm}
\usepackage{color}
\usepackage{changes}
\usepackage{balance}
\hypersetup{hidelinks}
\usepackage{url}
\usepackage{fancyhdr}

\algdef{SE}[DOWHILE]{Do}{doWhile}{\algorithmicdo}[1]{\algorithmicwhile\ #1}%

\newcommand{\mathactivatecomma}{%
  \begingroup\lccode`~=`\,
  \lowercase{\endgroup\edef~}{\mathchar\the\mathcode`\,\penalty0 }}

\algnewcommand{\Initialize}[1]{%
  \State \textbf{Initialize: $i \in \mathcal{V}$}
  \Statex \hspace*{\algorithmicindent}\parbox[t]{.8\linewidth}{\raggedright #1}
}

\algnewcommand{\Iteration}[1]{%
  \State \textbf{Iteration $(k\geq 0)$: $i \in \mathcal{V}$}
  \Statex \hspace*{\algorithmicindent}\parbox[t]{.8\linewidth}{\raggedright #1}
}

\algnewcommand{\Output}[1]{
  \State \textbf{Output: $i \in \mathcal{V}$}
  \Statex \hspace*{\algorithmicindent}\parbox[t]{.8\linewidth}{\raggedright #1}
}

\title{Unsupervised Light Field Depth Estimation via Multi-view Feature Matching with Occlusion Prediction}

\author{Shansi Zhang,~\IEEEmembership{Graduate Student Member,~IEEE}, Nan Meng,~\IEEEmembership{Member,~IEEE}, and Edmund Y. Lam,~\IEEEmembership{Fellow,~IEEE}

\thanks{The work is supported in part by the Research Grants Council of Hong Kong (GRF 17201822) and by ACCESS --- AI Chip Center for Emerging Smart Systems, Hong Kong SAR. (\textit{Corresponding author: Edmund Y. Lam})}
\thanks{Shansi Zhang and Edmund Y. Lam are with the Department of Electrical and Electronic Engineering, The University of Hong Kong, Hong Kong SAR, China (e-mail: sszhang@eee.hku.hk; elam@eee.hku.hk).}
\thanks{Nan Meng is with the Li Ka Shing Faculty of Medicine, The University of Hong Kong, Hong Kong SAR, China (e-mail: nanmeng@hku.hk).}
}

\usepackage{color}
\definecolor{darkspringgreen}{rgb}{0.09, 0.45, 0.27}

\begin{document}

\maketitle
\renewcommand{\headrulewidth}{0mm}

\pagestyle{empty}

\begin{abstract}
Depth estimation from light field (LF) images is a fundamental step for numerous applications. Recently, learning-based methods have achieved higher accuracy and efficiency than the traditional methods. However, it is costly to obtain sufficient depth labels for supervised training. In this paper, we propose an unsupervised framework to estimate depth from LF images. First, we design a disparity estimation network (DispNet) with a coarse-to-fine structure to predict disparity maps from different view combinations. It explicitly performs multi-view feature matching to learn the correspondences effectively. As occlusions may cause the violation of photo-consistency, we introduce an occlusion prediction network (OccNet) to predict the occlusion maps, which are used as the element-wise weights of photometric loss to solve the occlusion issue and assist the disparity learning. With the disparity maps estimated by multiple input combinations, we then propose a disparity fusion strategy based on the estimated errors with effective occlusion handling to obtain the final disparity map with higher accuracy. Experimental results demonstrate that our method achieves superior performance on both the dense and sparse LF images, and also shows better robustness and generalization on the real-world LF images compared to the other methods. 

\end{abstract}

\begin{IEEEkeywords}
Light field, unsupervised depth estimation, feature matching, occlusion prediction.
\end{IEEEkeywords}

\section{Introduction}\label{sec:introduction}
A light field (LF) camera can capture both the intensities and directions of the light rays~\cite{Ng2005,Lam2015} to obtain LF images, each of which consists of an array of sub-aperture images (SAIs) to record the scene from multiple viewpoints~\cite{Zhang2022,Zhang2023}. As the LF images contain rich geometric information, useful clues are available for depth (disparity) estimation. Usually, depth estimation is an essential task for scene understanding and also a crucial step for various LF applications and researches, such as auto refocusing~\cite{Fiss2014}, scene reconstruction~\cite{Kim2013}, novel view synthesis~\cite{Jin2020,Meng2021}, compressed sensing~\cite{Chen2017b}, and semantic segmentation~\cite{Sheng2022}. 

Traditional methods for LF depth estimation mainly adopt two approaches. The first approach is to explore the structure of epipolar-plane images (EPIs)~\cite{Wanner2014,Zhang2016,Zhang2017,Sheng2018}, where the pixels corresponding to the same scene point in different views form a line with a slope proportional to the disparity value. Another approach leverages the classical stereo matching to find the corresponding pixels among different views~\cite{Jeon2015,Tao2015,Lee2017,Huang2019}. However, EPI-based methods are mainly applicable to the densely sampled LF images, and the traditional stereo matching-based methods usually suffer from heavy computational costs. Recently, many learning-based methods~\cite{Sun2016,Heber2017,Shin2018,Shi2019,Tsai2020,Chen2021} have been proposed for LF depth estimation with improved accuracy and efficiency. They use a deep neural network to represent the estimator, which learns from the depth labels. However, it is costly to acquire sufficient, accurate depth annotations, especially for the real-world LF images, and the lack of training data usually leads to limited generalization ability. To alleviate the reliance on a large number of labeled data, some unsupervised learning-based methods~\cite{Peng2018,Zhou2020,Jin2022,Iwatsuki2022,Lin2022} are developed, which implicitly learn the correspondences by a plain network with the photo-consistency constraint~\cite{Tran2021} to minimize the warping errors. However, these methods are mainly applicable to the dense LF images but are not effective enough for the sparse LF images with large disparity values.

Our target is to develop an unsupervised LF depth framework applicable to both the dense and sparse LF images. We first design a disparity estimation network (DispNet) that explicitly performs multi-view feature matching by constructing memory-efficient cost volumes to learn the correspondences among the input views with a variety of disparity ranges. Our DispNet leverages a coarse-to-fine structure, with a coarse branch to estimate a initial disparity map and a refinement branch to further improve the disparity accuracy. Moreover, occlusion is a challenging issue in many LF tasks~\cite{Meng2019,Meng2021,Wang2023}, and it leads to the violation of photo-consistency in LF depth estimation. To tackle this issue, we introduce an occlusion prediction network (OccNet) during training to predict the occlusion maps, which are used as the element-wise weights of the photometric loss to mitigate the effect of occlusions on the disparity learning. In order to fully utilize the views of each LF image, multiple view combinations are input to the DispNet to obtain multiple estimated disparity maps. We then propose a disparity fusion strategy with effective occlusion handling to obtain the final disparity map with higher accuracy. The main contributions of our work are summarized as follows: 
\begin{itemize}
\item We develop a DispNet, which employs a coarse-to-fine structure and performs multi-view feature matching to estimate disparity maps from different input combinations. 
\item To tackle the occlusion issue, we introduce an OccNet for occlusion prediction, which aims to eliminate the adverse impact of occlusion on the training of DispNet. 
\item With the multiple estimated disparity maps, we propose a disparity fusion strategy with occlusion handling to obtain the final disparity map.
\item Experimental results demonstrate that our method achieves superior performance on both the dense and sparse LF images with better robustness and generalization compared to the other methods.
\end{itemize}


\section{Related Work}\label{sec:related work}
The existing methods for LF depth estimation are reviewed in terms of the traditional methods and the learning-based methods.

\subsection{Traditional Methods}
Traditional methods for LF depth estimation mainly focus on the EPI structure and stereo matching. For the EPI-based methods, Wanner and Goldluecke~\cite{Wanner2014} estimated the disparity maps locally by using EPI analysis, which works fast without the expensive matching cost minimization. Zhang et al.~\cite{Zhang2016} proposed a spinning parallelogram operator (SPO) to locate the lines in EPI and calculate their slopes to acquire the depth information. Zhang et al.~\cite{Zhang2017} exploited the line structure of EPI and the locally linear embedding (LLE) to estimate the local depth by minimizing the matching cost. Sheng et al.~\cite{Sheng2018} developed a strategy to extract EPIs in multiple directions besides the horizontal and vertical EPIs to calculate the local depth, which is combined with the predicted occlusion boundaries to obtain the final depth map. 

For the stereo matching-based methods, Jeon et al.~\cite{Jeon2015} constructed a cost volume to estimate the multi-view stereo correspondences, which was used to optimize the depths in weak texture regions. Then, the local depth map was refined iteratively by fitting the local quadratic function. Tao et al.~\cite{Tao2015} leveraged the shading information to improve the local shape estimation from defocus and correspondence, and developed a framework that exploits LF angular coherence for depth and shading optimization. Lee et al.~\cite{Lee2017} computed binary maps through foreground–background separation to obtain the disparity maps. Huang et al.~\cite{Huang2019} proposed a stereo matching algorithm using an empirical Bayesian framework, which employs the pseudo-random field to explore the statistical cues of LF. Zhang et al.~\cite{Zhang2020} leveraged graph spectral analysis to exploit the angular and spatial structure information for depth estimation.

Occlusion issue is often encountered in LF depth estimation since the photo-consistency assumption does not hold in the occlusion regions. There are some methods focusing on tackling this issue. Wang et al.~\cite{Wang2016} found that the photo-consistency still holds in about half of the views when the occlusions exist, and they predicted the occlusions, which was used as a regularizer to improve depth estimation. Williem et al.~\cite{Williem2016} focused on the robust depth estimation from noisy LF with occlusions. They introduced two data costs with angular entropy metric and adaptive defocus response to handle the occlusions and noises. Chen et al.~\cite{Chen2018} proposed a method with partially occluded region detection through super-pixel regularization and showed that even a simple least square model can achieve superior depth estimation after manipulating the label confidence and edge strength. 

These traditional methods usually involve complex optimization process and long execution time, and cannot achieve a good balance between the accuracy and efficiency. 

\subsection{Learning-based Methods}
The recent work on LF depth estimation mainly focuses on the learning-based methods, which usually achieves higher accuracy and inference efficiency. Most existing learning-based methods adopt supervised training using depth labels. Sun et al.~\cite{Sun2016} developed a convolutional neural network (CNN) to estimate LF disparity by extracting enhanced EPI features. Heber et al.~\cite{Heber2017} proposed a U-shaped network to extract LF geometric information, with 3D convolutional layers to examine the EPI volumes for robust depth prediction. Shin et al.~\cite{Shin2018} developed a CNN framework by taking as input the views from different angular dimensions. They also proposed some data augmentation methods for LF to overcome the deficiency of training data. Shi et al.~\cite{Shi2019} proposed a framework to learn depth from the dense and sparse LF images with three steps, including initial depth estimation by a fine-tuned network, occlusion-aware depth fusion and refinement by an additional network. Tsai et al.~\cite{Tsai2020} proposed a view selection network by learning an attention map to estimate the contribution of each view on depth. The attention map was constrained to be symmetric in accordance with the LF views. Chen et al.~\cite{Chen2021} developed a multi-level fusion network, which contains four branches to perform intra-branch and inter-branch fusion, and incorporates attention to select the features that can provide more useful information for depth. Wang et al.~\cite{Wang2022} proposed a fast approach to construct matching cost for LF depth estimation, which does not require any shifting operation and can also handle the occlusions. These supervised methods heavily rely on the labeled data, which results in poor generalization ability when the labeled data are not sufficient. 

Unsupervised methods can overcome the reliance on the labeled data. Peng et al.~\cite{Peng2018} proposed an unsupervised CNN framework by designing a combined loss with compliance and divergence constraints to estimate LF disparity. Zhou et al.~\cite{Zhou2020} developed an unsupervised monocular LF depth network, which was trained by the improved photometric losses and takes only one view as input. Jin et al.~\cite{Jin2022} proposed an unsupervised occlusion-aware framework by exploring the angular coherence among different LF subsets. Iwatsuki et al.~\cite{Iwatsuki2022} developed an unsupervised learning framework with pixel-wise weights to evaluate the warping errors and an edge loss to enforce edge alignment between the image and the disparity map. Lin et al.~\cite{Lin2022} proposed to integrate the traditional LF constraints into an unsupervised framework with an adaptive spatial-angular consistency loss. These methods directly output the disparity values from the last convolution layer without explicitly learning the correspondences among different views, which leads to poor performance when applied to the sparse LF images with large disparity values.
\begin{figure*}[!hbt]
\centering
\includegraphics[width=1\textwidth]{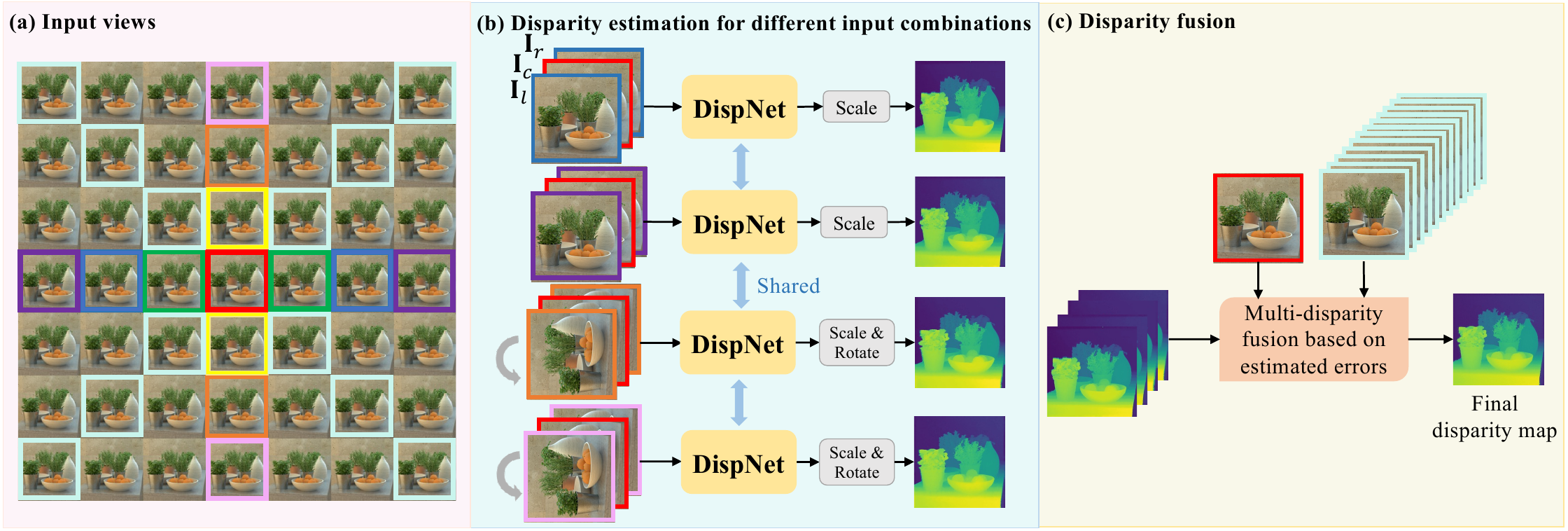} 
\caption{(a) The views input to the DispNet and used for determining the errors during fusion. (b) The DispNet takes as inputs the central view $\mathbf{I}_c$, the left source view $\mathbf{I}_{l}$ and the right source view $\mathbf{I}_{r}$. The views from the same column are rotated by $90^\circ$. Multiple disparity maps (after scaling and rotation) are obtained from different input combinations. (c) The estimated disparity maps are fused according to their estimated errors using the auxiliary views to obtain the final disparity map.}
\label{fig:inference}
\end{figure*}

\begin{figure}[!hbt]
\centering
\includegraphics[width=0.48\textwidth]{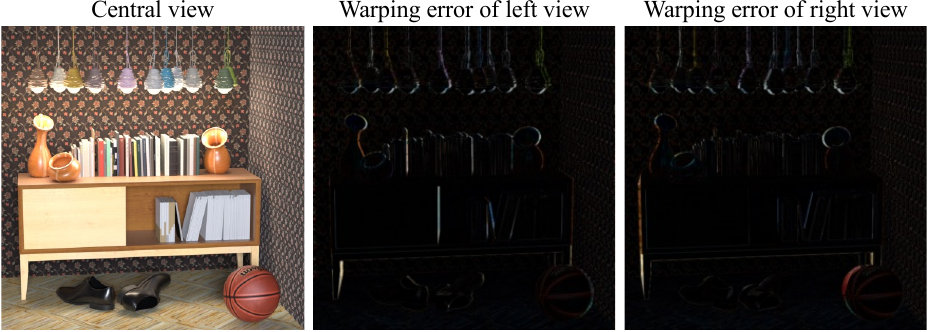} 
\caption{The central view and the warping error maps of the left and right views. The occlusion regions for the left and right views usually near the object boundaries and locate in the opposite positions.}
\label{fig:warp_error}
\end{figure}

\section{Proposed Method}\label{sec:method}
We first describe our overall framework for LF disparity estimation in Sec.~\ref{subsec:overall}. Then, we introduce the architecture of our DispNet in Sec.~\ref{subsec:disparity estimation}, the occlusion prediction method in Sec.~\ref{subsec:occlusion}, and the loss functions for training in Sec.~\ref{subsec:loss}. Finally, we introduce our disparity fusion strategy in Sec.~\ref{subsec:fusion}.

\subsection{Overall Framework}\label{subsec:overall}
A 4D LF image is represented as $\mathbf{L}\in \mathbb {R}^{U \times V \times X \times Y}$, with angular index $(u,v)$ and spatial index $(x,y)$. It consists of $U\times V$ SAIs, each of which records the scene from one viewpoint with a spatial resolution of $X\times Y$. Our target is to estimate the disparity map of the central view relative to its adjacent views. According to the LF geometry, the relationship between the central view and any other view in terms of the central disparity $\mathbf{d}(x,y)$ is expressed as 
\begin{align}\label{eq:warp}
\mathbf{L}(u_c,v_c,x,y)=\mathbf{L}\big(&u,v,x+\mathbf{d}(x,y)\times (u_c-u),\\ \nonumber
&y+\mathbf{d}(x,y)\times (v_c-v)\big),
\end{align}
where $(u_c,v_c)$ is the angular position of the central view.  

The overall framework of our method is depicted in Fig.~\ref{fig:inference}. Three views, including the central view $\mathbf{I}_c$, the left source view $\mathbf{I}_{l}$ and the right source $\mathbf{I}_{r}$ view (``source'' means that they are warped according to the disparity to reconstruct the central view), are fed to the DispNet. They are in the same row or column, and the two source views framed by the same color are located symmetrically to the central view. This input strategy can avoid the matching ambiguity caused by the occlusion without incorporating any redundant inputs, since each scene point is usually visible in at least two views of the three input views. Fig.~\ref{fig:warp_error} gives an intuitive illustration. The left view and right view are warped to the central view using the ground-truth disparity map of the central view, and their warping error maps are presented. The bright regions with large errors in the warping error maps correspond to the occlusion regions, which are visible in the central view but invisible in the left or right views. It can be seen that the occlusion regions for the left and right views usually near the object boundaries and locate in the opposite positions, which indicates that the pixel in the central view has correspondence in at least one view of the left and right views to enable accurate disparity estimation.

For a $7\times 7$ LF image (Fig.~\ref{fig:inference}(a)), there are totally $6$ input combinations. The views from the same column should be rotated by $90^{\circ}$ (counterclockwise) before being input to the network in order to convert the vertical disparity to be horizontal, and the corresponding output disparity map needs to be rotated by $-90^{\circ}$ (clockwise) to recover the orientation. In this way, only horizontal disparity estimation is involved to make the learning easier. Multiple disparity maps can be obtained from different input combinations by the shared DispNet, and they need to be scaled by the distance between the source views and the central view, as shown in Fig.~\ref{fig:inference}(b). Using nonadjacent views to estimate disparity helps to alleviate the inaccurate estimation caused by the narrow LF baseline. Then, these disparity maps are fused based on their estimated errors using the auxiliary views (in the diagonal direction) to obtain the final disparity map (Fig.~\ref{fig:inference}(c)). In what follows, we will introduce each part in detail.

\begin{figure*}[t]
\centering
\includegraphics[width=1\textwidth]{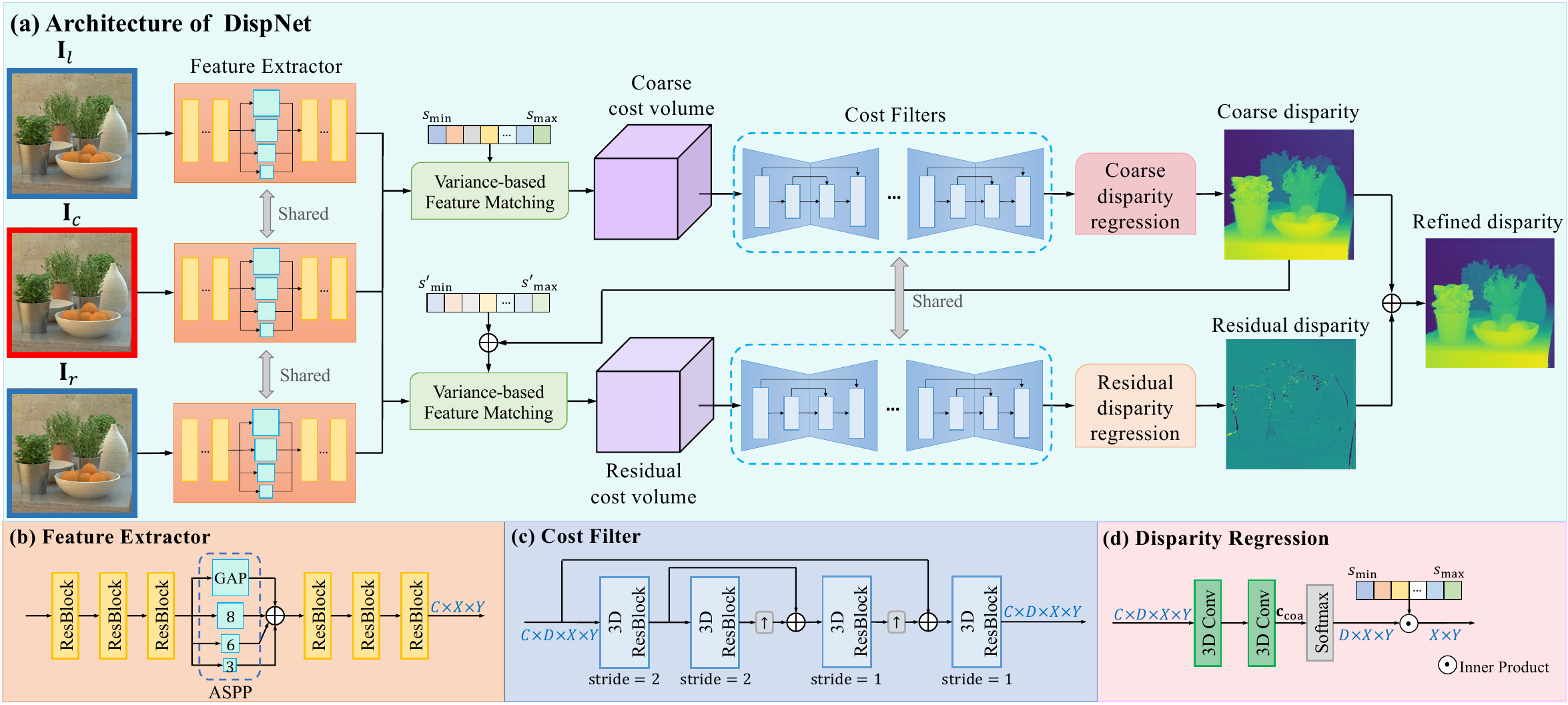} 
\caption{(a) Architecture of DispNet. It consists of two branches with shared feature extractor and cost filters to estimate the coarse disparity map and residual map by constructing the coarse and residual cost volumes with variance-based feature matching. (b) Feature extractor. (c) Cost filter. (d) Disparity regression.}
\label{fig:DispNet}
\end{figure*}

\begin{figure}[htb!]
\centering
\includegraphics[width=0.48\textwidth]{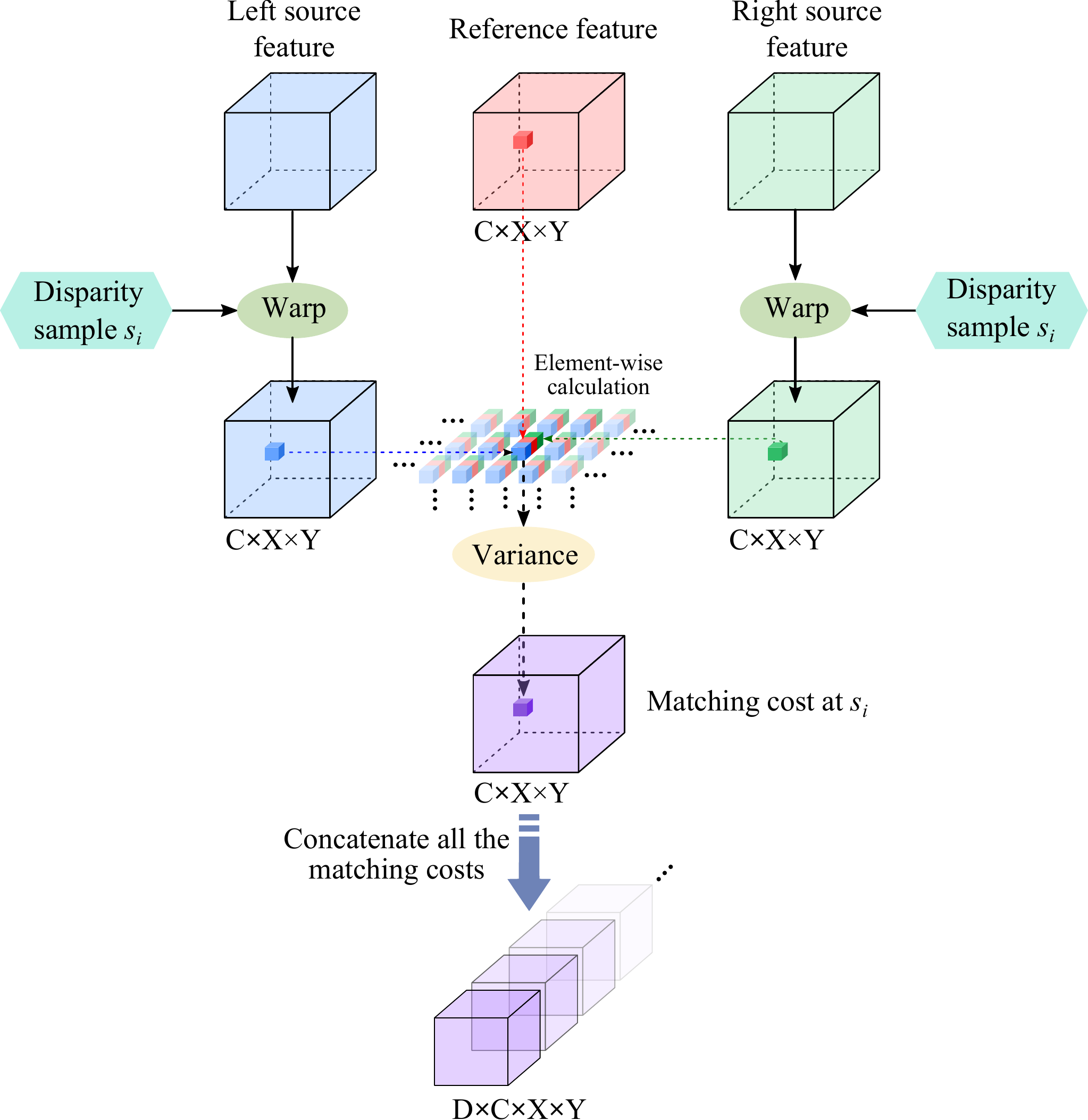} 
\caption{Variance-based feature matching. The left and right source features are warped according to a disparity sample $s_i$ to match the reference feature. The element-wise variance of the warped features and reference feature is calculated to obtain the matching cost at the disparity sample $s_i$. The final cost volume is obtained by concatenating the matching costs at all the disparity samples.}
\label{fig:cost_volume}
\end{figure}

\begin{figure*}[t]
\centering
\includegraphics[width=0.9\textwidth]{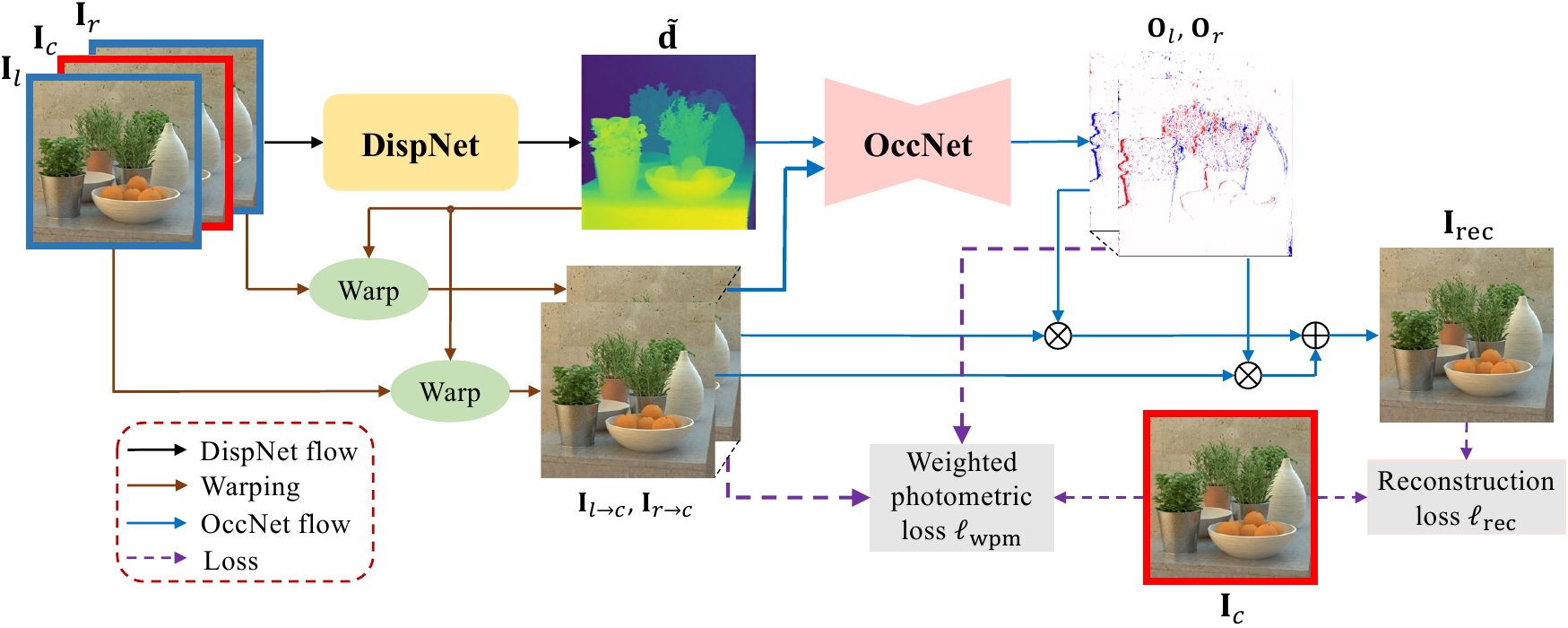} 
\caption{Training with predicted occlusion maps. The estimated disparity map from the DispNet is used to warp the two source views to the central view. Then, the concatenation of the warped views and the disparity map is input to the OccNet to predict the confidence maps for reconstructing the central view, which are treated as occlusion maps and used as the pixel-wise weights of the photometric loss $\ell_\mathrm{wpm}$. The OccNet is trained by the reconstruction loss $\ell_\mathrm{rec}$. The arrows for DispNet flow, warping, OccNet flow and loss are distinguished by different colors.}
\label{fig:training}
\end{figure*}

\subsection{Disparity Estimation}\label{subsec:disparity estimation}
The architecture of our DispNet is shown in Fig.~\ref{fig:DispNet}. The central view $\mathbf{I}_c$ and the two source views $\mathbf{I}_l$ and $\mathbf{I}_r$ are fed to a shared feature extractor. The feature extractor (Fig.~\ref{fig:DispNet}(b)) consists of several residual blocks, each of which has two $3\times 3$ convolution layers with leaky Rectified Linear Unit (ReLU) activation, and an atrous spatial pyramid pooling (ASPP)~\cite{Chen2017a} block to encode multi-scale features, which contains three dilated convolutions with rates $3$, $6$ and $8$ and a global average pooling (GAP) operation for global receptive field. 

The extracted features of the three views are used to construct the cost volumes through variance-based feature matching. The detailed procedures of constructing the coarse cost volume is illustrated in Fig.~\ref{fig:cost_volume}. First, we prescribe $D$ equally spaced disparity samples with the minimum value $s_\mathrm{min}$ and the maximum value $s_\mathrm{max}$, expressed by a vector $\mathbf{s}=[s_\mathrm{min},\cdots,s_i,\cdots,s_\mathrm{max}]^T$. Given the features of the three views with a size $C\times X\times Y$ ($C$ is the channel number), the left and right source features are warped to match the reference (central) feature by a disparity sample $s_i$ from $\mathbf{s}$. Then, the element-wise variance of the warped features and reference feature is calculated to measure the difference, which is treated as the matching cost at disparity sample $s_i$ (accurate disparity at a pixel should lead to a small variance). The cost volume is obtained by concatenating the matching costs at all the disparity samples, with a size $C\times D\times X\times Y$. Compared to the common approach used in stereo matching~\cite{Chang2018}, which builds the cost volume by concatenating the reference feature and warped source features, our variance-based feature matching is much more memory-efficient and can also adapt to any number of input views without increasing the size of cost volume. 

The coarse cost volume is further processed by the cost filters (Fig.~\ref{fig:DispNet}(c)), each of which consists of several 3D residual blocks with skip connections. The blocks with $\mathrm{stride}=2$ aim to downsample the cost volume to reduce memory consumption. Then, a coarse disparity regression module (Fig.~\ref{fig:DispNet}(d)) is employed, with 3D convolution layers to yield the coarse cost $\mathbf{c}_\mathrm{coa}\in \mathbb{R}^{D\times X\times Y}$, and the coarse disparity map $\tilde{\mathbf{d}}_\mathrm{coa}$ is obtained by
\begin{align}
\tilde{\mathbf{d}}_\mathrm{coa}(x,y)=\mathbf{s}\cdot \mathrm{softmax}\big(\mathbf{c}_\mathrm{coa}(x,y)\big),
\end{align}
where $\mathrm{softmax}(\cdot)$ is the softmax function to obtain the probability of each disparity sample, and $\cdot$ denotes the inner product.

To further refine the disparity map, a residual cost volume is constructed by using the same extracted features. Similarly, we set a residual sample vector $\mathbf{s'}=[s'_\mathrm{min},\cdots,s'_i,\cdots,s'_\mathrm{max}]^T$ with the minimum value $s'_\mathrm{min}$ and the maximum value $s'_\mathrm{max}$. The left and right source features are warped according to the coarse disparity map plus a residual sample $s'_i$ from $\mathbf{s'}$, with $\tilde{\mathbf{d}}_\mathrm{coa}(x,y)+s'_i$ at each position, to match the reference feature. The value and interval of residual sampling are much smaller than that of coarse sampling in order to improve the disparity accuracy. The matching cost at each residual sample is derived by calculating the variance of the warped and reference features, and the residual cost volume is obtained by concatenating the matching costs at all the residual samples. Then, the residual cost volume is processed by the shared cost filters and the residual disparity regression module to derive the residual map $\tilde{\mathbf{d}}_\mathrm{res}$, with
\begin{align}
\tilde{\mathbf{d}}_\mathrm{res}(x,y)=\mathbf{s'}\cdot \mathrm{softmax}\big(\mathbf{c}_\mathrm{res}(x,y)\big),
\end{align}
where $\mathbf{c}_\mathrm{res}\in\mathbb{R}^{D\times X\times Y}$ is the residual cost.

Thus, the refined disparity map $\tilde{\mathbf{d}}$ is obtained by
\begin{align}
\tilde{\mathbf{d}}=\tilde{\mathbf{d}}_\mathrm{coa}+\tilde{\mathbf{d}}_\mathrm{res}.
\end{align}

When the input views are not adjacent, the output disparity maps from DispNet need to be scaled according to the view distance. In addition, the disparity maps predicted by the views from the same column need to be rotated by $-90^\circ$ to recover the orientation. Thus, the estimated central disparity map $\hat{\mathbf{d}}$ is obtained by 

\begin{align}
\hat{\mathbf{d}}=
\begin{cases}
\frac{\tilde{\mathbf{d}}}{u_c-u_l}, & \text{from the same row} \\
\mathrm{rot}_{-90^\circ}(\frac{\tilde{\mathbf{d}}}{v_c-v_l}), & \text{from the same column}
\end{cases}
\end{align}
where $u_l$ and $v_l$ are the angular coordinates of the left source view, and $\mathrm{rot}_{-90^\circ}$ means rotating by $-90^\circ$.

\subsection{Occlusion Prediction}\label{subsec:occlusion}

During training, $\tilde{\mathbf{d}}$ is used to warp the left and right source views to the central view, yielding $\mathbf{I}_{l\rightarrow c}$ and $\mathbf{I}_{r\rightarrow c}$, with
\begin{align}
\mathbf{I}_{l\rightarrow c}(x,y)&=\mathbf{I}_{l}\big(x+\tilde{\mathbf{d}}(x,y),y\big),\\
\mathbf{I}_{r\rightarrow c}(x,y)&=\mathbf{I}_{r}\big(x-\tilde{\mathbf{d}}(x,y),y\big).
\end{align}

To predict the occlusion regions for the two source views, we introduce an OccNet that takes the concatenation of $\mathbf{I}_{l\rightarrow c}$, $\mathbf{I}_{r\rightarrow c}$ and $\tilde{\mathbf{d}}$ as inputs, as shown in Fig.~\ref{fig:training}. The OccNet adopts a U-shape structure with residual blocks and skip connections. A softmax activation is used in the last convolution layer to output the confidence maps, $\mathbf{O}_{l}$ and $\mathbf{O}_{r}$ for $\mathbf{I}_{l\rightarrow c}$ and $\mathbf{I}_{r\rightarrow c}$, respectively, which are used to reconstruct the central view, with,
\begin{align}
\mathbf{I}_\mathrm{rec}=\mathbf{O}_{l}\odot \mathbf{I}_{l\rightarrow c}+\mathbf{O}_{r}\odot \mathbf{I}_{r\rightarrow c},
\end{align}
where $\mathbf{I}_\mathrm{rec}$ is the reconstructed central view, $\odot$ represents the element-wise multiplication, and $\mathbf{O}_{l}(x,y)+\mathbf{O}_{r}(x,y)=1$. The OccNet is trained by the reconstruction loss, with
\begin{align}\label{eq:reconstruction loss}
\ell_\mathrm{rec}=\|\mathbf{I}_\mathrm{rec}-\mathbf{I}_c\|_1,
\end{align}
where $\|\cdot\|_1$ is the $\ell_1$-norm operator. 

The photometric loss for unsupervised disparity training is based on the photo-consistency assumption, which does not hold in the occlusion regions. Therefore, the pixel-wise weights of the photometric loss in the occlusion regions are expected to be small. The confidence maps $\mathbf{O}_{l}$ and $\mathbf{O}_{r}$ can be treated as the occlusion maps of the two warped views, since the occlusion regions usually lead to relatively larger warping errors, and therefore less confidence for the reconstruction. Thus, we propose the weighted photometric loss with the occlusion maps, expressed as
\begin{align}\label{eq:wpm loss}
\ell_\mathrm{wpm}&=\frac{1}{XY}\sum_{x,y}\mathbf{O}_{l}^{(x,y)}\odot|\mathbf{I}_{l\rightarrow c}^{(x,y)}-\mathbf{I}_c^{(x,y)}|\\ \nonumber
&\quad+\frac{1}{XY}\sum_{x,y}\mathbf{O}_{r}^{(x,y)}\odot|\mathbf{I}_{r\rightarrow c}^{(x,y)}-\mathbf{I}_c^{(x,y)}|.
\end{align}

The OccNet plays an important role during the training of DispNet. First, it helps to enforce the similarity between the warped source views and the central view for disparity learning through the reconstruction loss, as it is jointly trained with the DispNet. Second, with the improvement of estimated disparity map, large warping errors mainly lie in the occlusion regions, which can be identified by the OccNet to alleviate the adverse impact on further improvement. Note that the OccNet is only employed during training to address the occlusion issue and assist the disparity learning without influencing the inference efficiency.

\subsection{Loss Function}\label{subsec:loss}
In addition to the weighted photometric loss $\ell_\mathrm{wpm}$ and the reconstruction loss $\ell_\mathrm{rec}$, we apply a structural similarity (SSIM) loss~\cite{Zhang2021a} to further enforce the similarity, with
\begin{align}
\ell_\mathrm{SSIM}=1-\frac{\mathrm{SSIM}(\mathbf{I}_{l\rightarrow c},\mathbf{I}_c)+\mathrm{SSIM}(\mathbf{I}_{r\rightarrow c},\mathbf{I}_c)}{2}.
\end{align}

To improve the smoothness of the estimated disparity map while preserving the boundary structures of the objects, we leverage the structure-aware smoothness loss~\cite{Godard2017,Zhang2021b}, expressed as
\begin{align}
\ell_\mathrm{smd}=\frac{1}{XY}\sum_{x,y}|\nabla \tilde{\mathbf{d}}^{(x,y)}|\odot \exp{(-\eta |\nabla \mathbf{I}_c^{(x,y)}|)},
\end{align}
where $\nabla$ denotes the gradients along both the horizontal and vertical directions, and $\eta$ is a hyperparameter for structure preservation. 

Moreover, we apply a similar smoothness loss to the occlusion map, with
\begin{align}
\ell_\mathrm{smo}=\frac{1}{XY}\sum_{x,y}|\nabla \mathbf{O}_{l}^{(x,y)}|\odot \exp{(-\eta |\nabla \mathbf{I}_c^{(x,y)}|)},
\end{align}
which is only applied to $\mathbf{O}_{l}$ since $\mathbf{O}_{r}=\mathbf{1}-\mathbf{O}_{l}$.

The DispNet and OccNet are trained simultaneously by the following full loss,
\begin{align}\label{eq:full loss}
\ell_\mathrm{full}=\ell_\mathrm{wpm}+\ell_\mathrm{rec}+\alpha_1\ell_\mathrm{SSIM}+\alpha_2\ell_\mathrm{smd}+\alpha_3\ell_\mathrm{smo},
\end{align}
where $\ell_\mathrm{wpm}$ and $\ell_\mathrm{rec}$ are the necessary losses, and the others are the auxiliary losses with coefficients $\alpha_1\sim \alpha_3$. Note that these loss terms are also applied to the coarse disparity map $\tilde{\mathbf{d}}_\mathrm{coa}$, but we omit the procedures for simplicity.  

\subsection{Multi-disparity Fusion Based on Estimated Errors}\label{subsec:fusion}
Multiple disparity maps can be obtained by the DispNet with different input combinations. To obtain the final disparity map, we propose a disparity fusion strategy to merge these disparity maps. 

Suppose that we have $n$ estimated disparity maps $\{\hat{\mathbf{d}}_j\}_{j=1}^{n}$ and $Z$ auxiliary views to evaluate their accuracy. With each disparity map, the auxiliary views are warped to the central view, and the warping errors are calculated. However, some of the warping errors are not accurate due to the occlusions. If occlusions exist in some of the auxiliary views, the warping errors at the corresponding positions would be large, resulting in large variances among the $Z$ warping errors. Thus, we calculate the standard deviation of the warping errors to judge if occlusion exists at each position, and obtain a binary mask. Here, we define $\bm{\epsilon}_{j}\in\mathbb{R}^{X\times Y\times Z}$, indexed by $(x,y,z)$, as the warping error maps of the $Z$ auxiliary views using the disparity map $\hat{\mathbf{d}}_j$, and $\sigma_z(\cdot)$ as the standard deviation along $Z$ dimension. Then, a binary mask $\mathbf{M}_j$ is formulated as
\begin{align}\label{eq:M}
\mathbf{M}_j^{(x,y)}=
\begin{cases}
1,& \sigma_z(\bm{\epsilon}_{j}^{(x,y,z)})>\theta(q)\\
0,& \text{otherwise}
\end{cases}
\end{align}
where $\theta(\cdot)$ is to determine the threshold for occlusion using the quantile $q$ of $\sigma_z(\bm{\epsilon}_{j}^{(x,y,z)})$. The effect of $q$ is analyzed in Sec.~\ref{sec:fusion}.

For a pixel with occlusion, we use the median value of the warping errors to represent its estimated error, which can eliminate the influence of large warping errors due to occlusion. Otherwise, the estimated error is represented by the mean of all the warping errors. Then, the estimated error map $\mathbf{e}_{j}$ for $\hat{\mathbf{d}}_j$ is obtained by
\begin{equation}
\mathbf{e}_{j}^{(x,y)}=\mathrm{median}_z\big(\bm{\epsilon}_{j}^{(x,y,z)}\big)\odot \mathbf{M}_j +\mathrm{mean}_z\big(\bm{\epsilon}_{j}^{(x,y,z)}\big)\odot(\mathbf{1}-\mathbf{M}_j),
\end{equation}
where $\mathrm{median}_z(\cdot)$ and $\mathrm{mean}_z(\cdot)$ denote the median and mean along $Z$ dimension.

With the above procedures, we can derive the error maps $\{\mathbf{e}_j\}_{j=1}^{n}$ for all the disparity maps. Next, we need to merge these disparity maps according to their estimated errors. Here, we consider several different fusion approaches. The first one is the minimum error fusion by choosing the pixels with the minimum errors from each disparity map, and the final disparity map $\hat{\mathbf{d}}_\mathrm{final}$ is derived by
\begin{align}
&j'=\arg\min_j\big(\mathbf{e}_j(x,y)\big), \\
&\hat{\mathbf{d}}_\mathrm{final}(x,y)=\hat{\mathbf{d}}_{j'}(x,y).
\end{align}

The second approach is the weighted fusion, where the weights are obtained by the softmax function and negatively correlated with the errors. Moreover, we can choose different numbers of disparity pixels for weighted fusion at each position. If $n'$ ($n' \le n$) disparity pixels with the smallest errors are used at each position, the final disparity map is obtained by
\begin{align}\label{eq:partial weighted fusion}
\mathbf{W}(x,y)&=\mathrm{softmax} \big(-\mathbf{E}(x,y)\big),\\ 
\hat{\mathbf{d}}_\mathrm{final}(x,y)&=\sum_{j=1}^{n'} \mathbf{w}_j(x,y)\times \hat{\mathbf{d}}_j(x,y),
\end{align}
where $\mathbf{E}(x,y)=\{\mathbf{e}_j(x,y)\}_{j=1}^{n'}$ denotes the smallest $n'$ errors at position $(x,y)$, and $\mathbf{W}(x,y)=\{\mathbf{w}_j(x,y)\}_{j=1}^{n'}$ denotes their corresponding weights.


\begin{figure*}[t]
\centering
\includegraphics[width=1\textwidth]{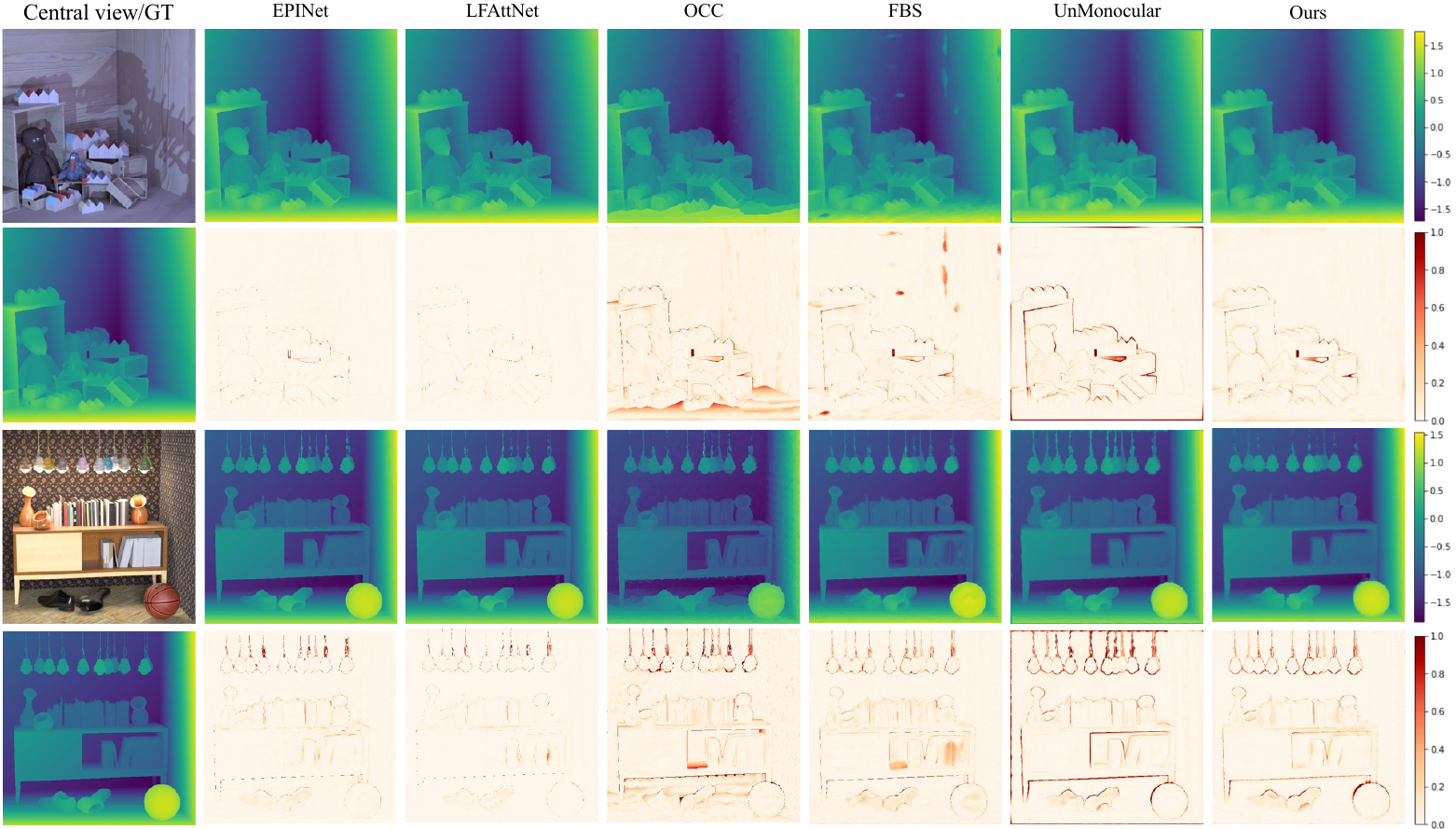} 
\caption{Visual results of different methods on the scenes from the HCI dataset~\cite{Honauer2016}, with the error maps relative to the ground truths.}
\label{fig:HCI comparison}
\end{figure*}

\begin{table*}[t]
\caption{Evaluation on the scenes from HCI dataset. \textbf{Bold}: Best among the unsupervised methods.}
\centering
\resizebox{\textwidth}{!}{
\begin{tabular}{c|cccc|cccc|cccc|cccc}
\toprule
\multirow{2}{*}{} & \multicolumn{4}{c|}{Dino} &\multicolumn{4}{c|}{Sideboard} &\multicolumn{4}{c|}{Backgammon} &\multicolumn{4}{c}{Pyramids}\\
\cmidrule{2-17}
Methods & MSE$\downarrow$ & BPR$\downarrow$ & BPR$\downarrow$ & BPR$\downarrow$ & MSE$\downarrow$ & BPR$\downarrow$ & BPR$\downarrow$ & BPR$\downarrow$ & MSE$\downarrow$ & BPR$\downarrow$ & BPR$\downarrow$ & BPR$\downarrow$ & MSE$\downarrow$ & BPR$\downarrow$ & BPR$\downarrow$ & BPR$\downarrow$ \\[0.5ex]
& ($\times100$) & (0.07) & (0.03) & (0.01) & ($\times100$) & (0.07) & (0.03) & (0.01) & ($\times100$) & (0.07) & (0.03) & (0.01) & ($\times100$) & (0.07) & (0.03) & (0.01)\\
\midrule
\textbf{Supervised} &&&&&&&&&&&&&&&&\\
EPINet~\cite{Shin2018}  & 0.167 & 1.286 & 3.452 & 22.401 & 0.742 & 4.277 & 10.824 & 37.999 & 3.629 & 3.580 & 6.289 & 20.899 & 0.008 & 0.192 & 0.913 & 11.876\\
LFAttNet~\cite{Tsai2020} & 0.093 & 0.848 & 2.340 & 12.224 & 0.531 & 2.870 & 7.243 & 20.739 & 3.648 & 3.126 & 3.984 & 11.582 &0.004 & 0.195 & 0.489 & 2.063\\
\midrule
\textbf{Non-learning} &&&&&&&&&&&&&&&&\\
OCC~\cite{Wang2016}  & 0.944 & 15.366 & 50.167 & 88.810 & 2.073 & 17.910 & 50.550 & 84.653 & 22.782 & 13.522 & 44.899 & 91.402 & 0.077 & 1.450 & 25.574 & 92.860\\
FBS~\cite{Lee2017}   & 0.664 & 8.427 & 23.533 & 65.390 & 1.072 & 13.296 & 32.516 & 70.042 & 5.805 & 10.162 & 22.181 & 65.407 & 0.029 & 0.549 & 5.705 & 78.243\\
\midrule
\textbf{Unsupervised} &&&&&&&&&&&&&&&&\\ 
UnCNN~\cite{Peng2018}  & 1.807 & 23.660 & 47.876 & 78.724 & 3.149 & 26.173 & 45.384 & 82.924 & 11.034 & 31.783 & 65.583 & 87.987 & 0.191 & 10.849 & 43.972 & 79.113\\
UnMonocular~\cite{Zhou2020} & 1.031 & \textbf{5.402} & \textbf{14.757} & \textbf{43.258} & 2.770 & \textbf{10.947} & \textbf{23.646} & 61.406 & 11.833 & 12.311 & 28.524 & 68.312 & 0.027 & \textbf{0.262} & 8.725 & 35.594\\
UnPlug~\cite{Iwatsuki2022} & 0.788 & 6.178 & - & - & 1.999 & 12.766 & - & - & 9.399 & 14.200 & - & - & \textbf{0.022} & 0.658 & - & -\\
UnOcc~\cite{Jin2022}  & \textbf{0.63} & 8.25 & - & - & 1.79 & 14.20 & - & - & 6.684 & 14.371 & - & - & 0.213 & 7.348 & - & -\\
Ours & 0.650 & 6.586 & 16.722 & 46.380 & \textbf{1.738} & 12.013 & 25.848 & \textbf{58.744} & \textbf{5.740} & \textbf{10.710} & \textbf{18.452} & \textbf{51.066} & 0.023 & 0.670 & \textbf{4.720} & \textbf{24.314}\\
\bottomrule
\end{tabular}}
\label{tab:HCI comparison}
\end{table*}

\begin{figure*}[t]
\centering
\includegraphics[width=1\textwidth]{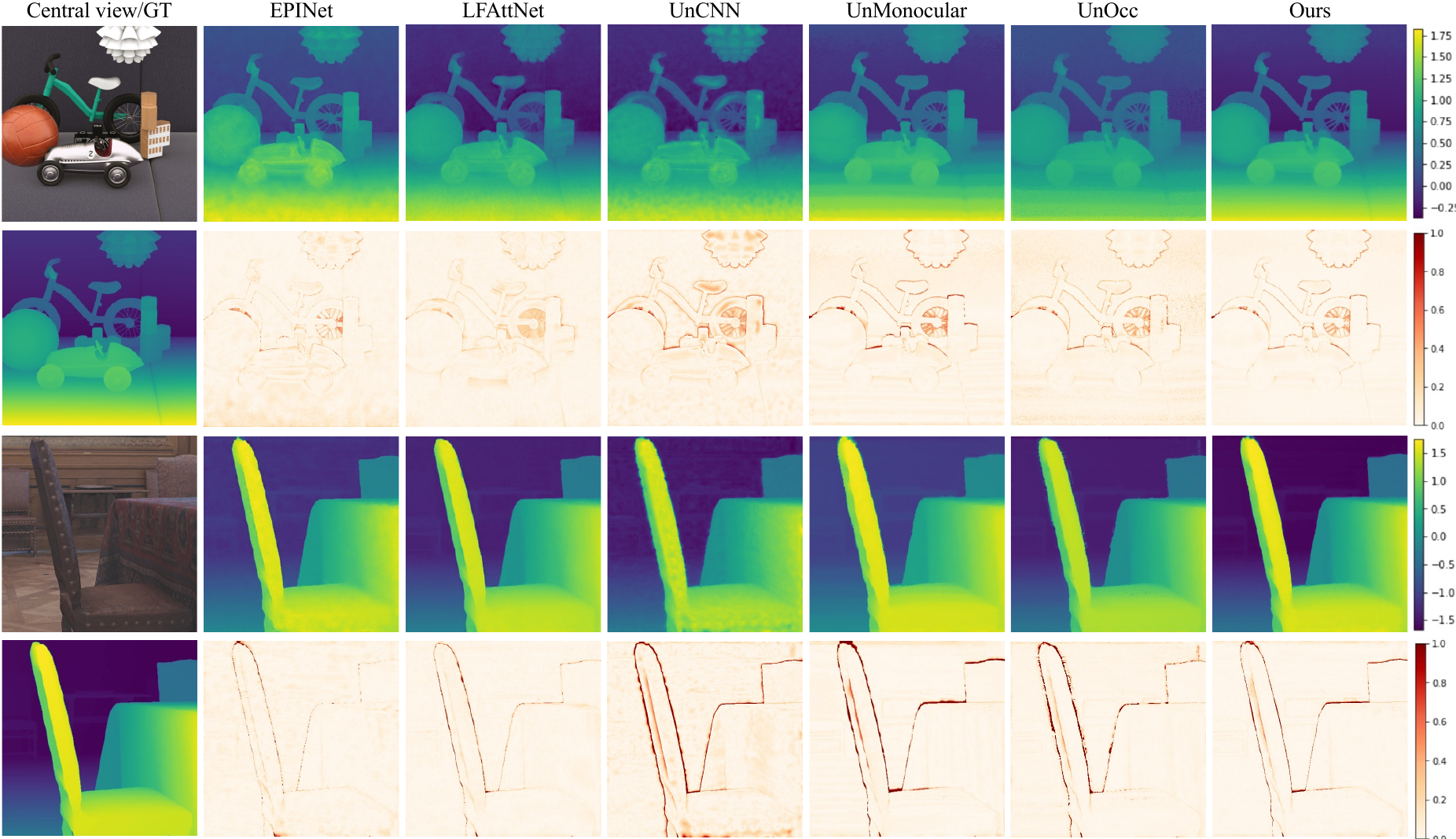} 
\caption{Visual results of different methods on the scenes from the DLF dataset~\cite{Shi2019}, with the error maps relative to the ground truths.}
\label{fig:DLF comparison}
\end{figure*}

\begin{table*}[t]
\caption{Evaluation on the scenes from DLF dataset. \textbf{Bold}: Best among the unsupervised methods.}
\centering
\resizebox{\textwidth}{!}{
\begin{tabular}{c|cccc|cccc|cccc|cccc}
\toprule
\multirow{2}{*}{} &\multicolumn{4}{c|}{Toys} &\multicolumn{4}{c|}{Antiques} &\multicolumn{4}{c|}{Pinenuts white} &\multicolumn{4}{c}{Smiling crowd roses}\\
\cmidrule{2-17}
Methods & MSE$\downarrow$ & BPR$\downarrow$ & BPR$\downarrow$ & BPR$\downarrow$ & MSE$\downarrow$ & BPR$\downarrow$ & BPR$\downarrow$ & BPR$\downarrow$ & MSE$\downarrow$ & BPR$\downarrow$ & BPR$\downarrow$ & BPR$\downarrow$ & MSE$\downarrow$ & BPR$\downarrow$ & BPR$\downarrow$ & BPR$\downarrow$\\[0.5ex]
& ($\times100$) & (0.07) & (0.03) & (0.01) & ($\times100$) & (0.07) & (0.03) & (0.01) & ($\times100$) & (0.07) & (0.03) & (0.01) & ($\times100$) & (0.07) & (0.03) & (0.01)\\
\midrule
\textbf{Supervised} &&&&&&&&&&&&&&&&\\
EPINet~\cite{Shin2018}  & 0.431 & 15.540  & 42.438 & 75.800 & 1.265 & 6.992 & 32.292 & 72.562 & 0.509 & 15.232 & 35.826 & 69.557 & 3.148 & 14.997 & 41.297 & 77.024\\
LFAttNet~\cite{Tsai2020}  & 0.405 & 10.231 & 35.711 & 74.166 & 0.827 & 4.206 & 21.862 & 67.076 & 0.406 & 10.698 & 27.042 & 65.995 & 2.025 & 12.454 & 33.356 & 66.702\\
\midrule
\textbf{Unsupervised} &&&&&&&&&&&&&&&&\\
UnCNN~\cite{Peng2018}  & 0.960 & 20.031 & 53.104 & 82.405 & 4.876 & 21.944 & 56.391 & 84.684 & 2.099 & 35.955 & 65.235 & 89.028 & 5.143 & 32.653 & 62.062 & 86.544\\
UnMonocular~\cite{Zhou2020} & 0.859 & 8.783 & 47.252 & 82.033 & 4.551 & 7.073 & 21.986 & 63.595 & 0.803 & 14.146 & 39.468 & 74.769 & 6.257 & 13.522 & 28.691 & 73.238\\
UnCon~\cite{Lin2022} & 0.886 & 12.238 & 57.656 & 84.407 & 3.322 & 10.859 & 43.703 & 80.262 & 0.540 & 14.402 & 54.013 & 84.173 & 3.916 & 14.595 & 31.483 & 72.385 \\
UnOcc\cite{Jin2022} & 1.021 & 18.497 & 44.714 & 76.093 & 4.034 & 7.018 & 22.748 & 61.508 & 0.706 & 21.065 & 49.413 & 79.825 & 3.678 & 15.403 & 32.678 & 71.964\\
Ours        & \textbf{0.643}  & \textbf{6.745} & \textbf{24.210} & \textbf{60.513} & \textbf{2.336} & \textbf{5.514} & \textbf{12.937} & \textbf{42.088} & \textbf{0.434} & \textbf{8.106} & \textbf{25.164} & \textbf{61.817} & \textbf{3.243} & \textbf{11.325} & \textbf{23.254} & \textbf{50.507}\\
\bottomrule

\end{tabular}}
\label{tab:DLF comparison}
\end{table*}

\begin{figure*}[t]
\centering
\includegraphics[width=1\textwidth]{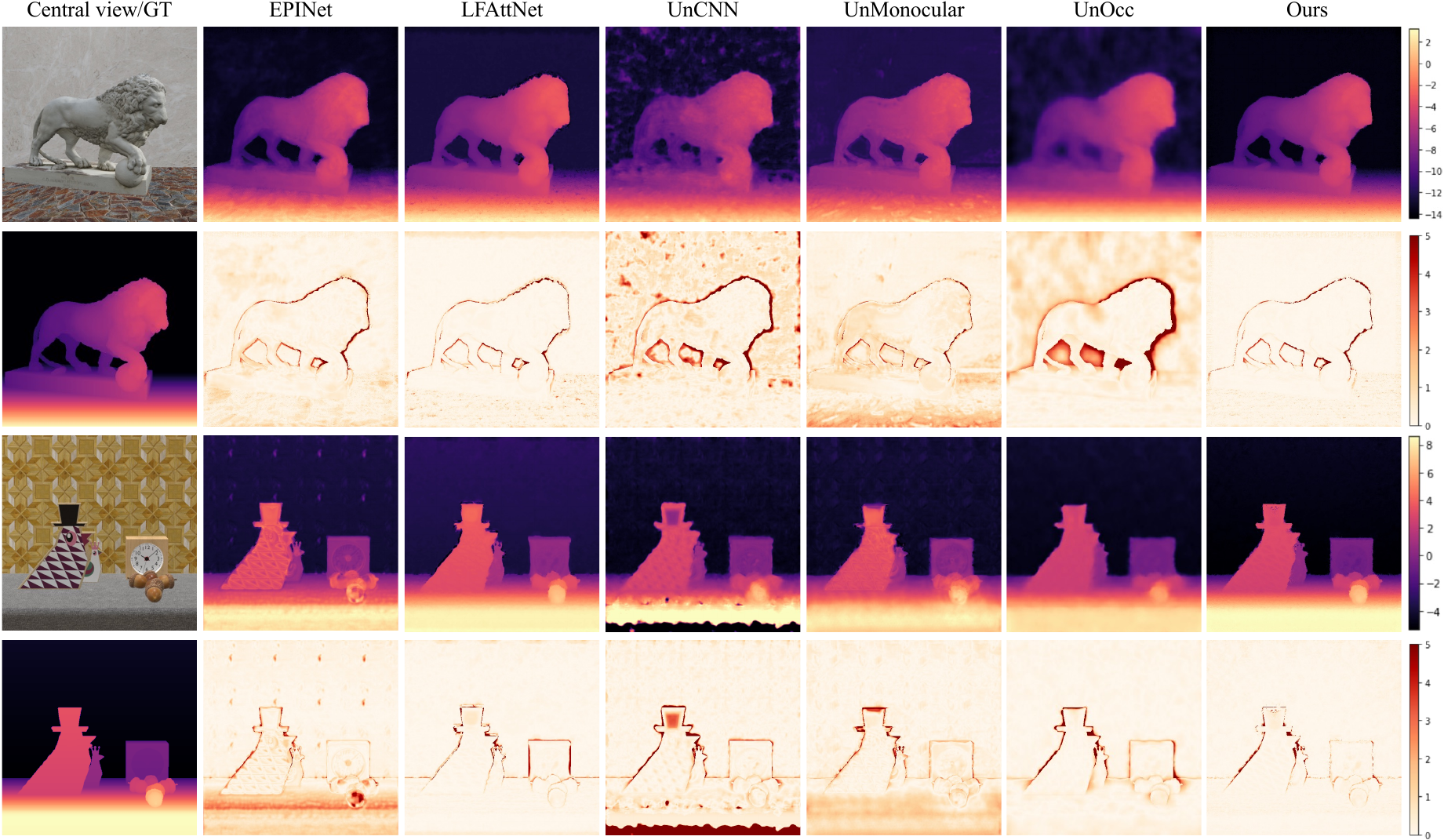} 
\caption{Visual results of different methods on the sparse LF images from the SLF dataset~\cite{Shi2019}, with the error maps relative to the ground truths.}
\label{fig:SLF comparison}
\end{figure*}

\begin{table*}[t]
\caption{Evaluation on the scenes from SLF dataset. \textbf{Bold}: Best among the unsupervised methods.}
\centering
\resizebox{\textwidth}{!}{
\begin{tabular}{c|cccc|cccc|cccc|cccc}
\toprule
\multirow{2}{*}{} &\multicolumn{4}{c|}{Lion} &\multicolumn{4}{c|}{Rooster clock} &\multicolumn{4}{c|}{Toy bricks} &\multicolumn{4}{c}{Electro devices}\\
\cmidrule{2-17}
Methods & MSE$\downarrow$ & BPR$\downarrow$ & BPR$\downarrow$ & BPR$\downarrow$ & MSE$\downarrow$ & BPR$\downarrow$ & BPR$\downarrow$ & BPR$\downarrow$ & MSE$\downarrow$ & BPR$\downarrow$ & BPR$\downarrow$ & BPR$\downarrow$ & MSE$\downarrow$ & BPR$\downarrow$ & BPR$\downarrow$ & BPR$\downarrow$ \\[0.5ex]
&  & (0.3) & (0.1) & (0.05) &  & (0.3) & (0.1) & (0.05) &  & (0.3) & (0.1) & (0.05) &  & (0.3) & (0.1) & (0.05)\\
\midrule
\textbf{Supervised} &&&&&&&&&&&&&&&& \\
 EPINet~\cite{Shin2018}  & 0.476 & 31.457 & 68.138 & 82.954 & 0.740 & 36.868 & 71.562 & 85.147 & 1.057 & 34.658 & 61.957 & 77.491 & 1.112 & 39.809 & 64.455 & 81.793\\
                          LFAttNet~\cite{Tsai2020}  & 0.372 & 12.994 & 29.594 & 49.327 & 0.278 & 6.831 & 25.813 & 50.920 & 0.738 & 11.607 & 29.018 & 52.576 & 0.703 & 13.819 & 38.731 & 60.993\\
\midrule
\textbf{Unsupervised} &&&&&&&&&&&&&&&& \\
 UnCNN~\cite{Peng2018}  & 1.442 & 58.221 & 84.867 & 92.373 & 9.213 & 34.586 & 75.021 & 87.383 & 2.254 & 53.922 & 82.172 & 90.950 & 2.181 & 41.163 & 72.970 & 85.866\\
                              UnMonocular~\cite{Zhou2020} & 0.760 & 48.305 & 87.371 & 94.673 & 0.664 & 39.931 & 73.601 & 86.370 & 1.226 & 47.594 & 71.098 & 84.982 & 2.736 & 45.612 & 75.238 & 87.332\\
                              UnCon~\cite{Lin2022} & 1.192 & 35.493 & 75.392 & 87.891 & 0.491 & 16.928 & 53.889 & 75.597 & 1.741 & 50.471 & 79.470 & 89.474 & 3.604 & 42.443 & 75.654 & 87.271 \\
                              UnOcc\cite{Jin2022} & 1.454 & 53.460 & 80.650 & 90.212 & 0.421 & 11.148 & 42.627 & 68.151 & 1.906 & 61.457 & 84.587 & 91.570 & 2.720 & 40.834 & 71.622 & 84.935 \\
                              Ours       & \textbf{0.360} & \textbf{8.766} & \textbf{24.420} & \textbf{47.911} & \textbf{0.261} & \textbf{5.796} & \textbf{25.303} & \textbf{48.668} & \textbf{0.772} & \textbf{10.048} & \textbf{30.427} & \textbf{55.732} & \textbf{0.842} & \textbf{13.631} & \textbf{36.038} & \textbf{57.880}\\
\bottomrule

\end{tabular}}
\label{tab:SLF comparison}
\end{table*}

\begin{figure*}[!hbt]
\centering
\includegraphics[width=1\textwidth]{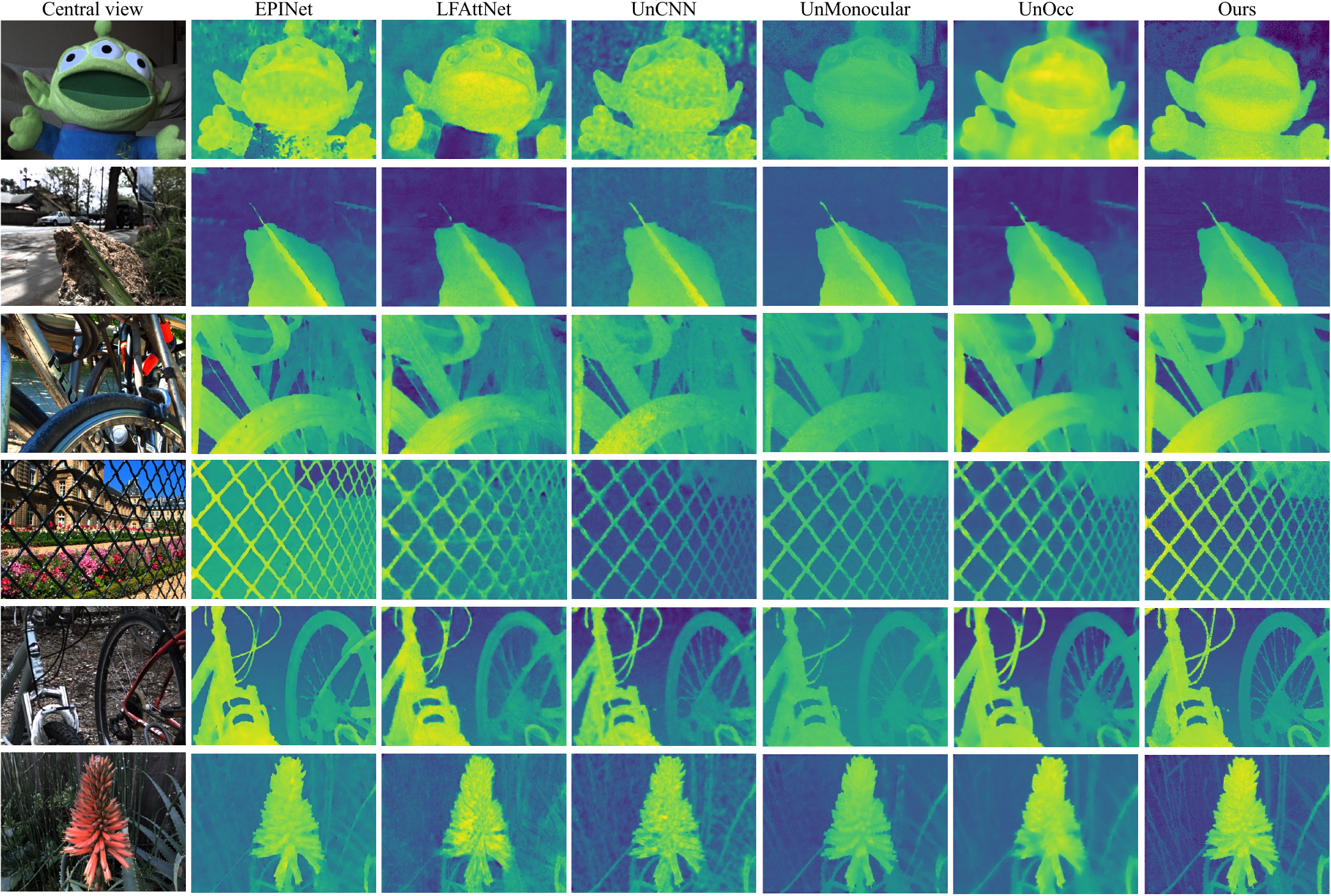} 
\caption{Visual results of different methods on the real-world LF images from the datasets~\cite{Raj2016,Kalantari2016,Rerabek2016}. Zoom in for best view.}
\label{fig:Real comparison}
\end{figure*}

\section{Experiments}\label{sec:experiments}

\subsection{Datasets}
To evaluate the model performance comprehensively, we used both the synthetic and real-world LF datasets. The synthetic LF datasets include both the densely and sparsely sampled LF images. The synthetic dense LF images are from the HCI dataset~\cite{Honauer2016} and the DLF dataset~\cite{Shi2019}, with a disparity range $[-4,4]$ pixels between the adjacent views. The synthetic sparse LF images are from the SLF dataset~\cite{Shi2019}, with a much larger disparity range $[-20,20]$ pixels. The real-world LF images are from the Stanford Lytro LF Archive~\cite{Raj2016}, Kalantari~\cite{Kalantari2016} and EPFL LF~\cite{Rerabek2016} datasets. They can also be treated as the dense LFs since the disparity values between the adjacent views are very small. For all the LF images, the central $7\times 7$ SAIs were used to estimate the disparity maps for the central views. 

\subsection{Implementation Details}
Regarding the DispNet, the maximum channel number within the feature extractor is $128$, and the number of cost filters is $2$. For the dense LFs, The range of disparity samples was set to $[-12,12]$ (three times the disparity range $[-4,4]$) with a coarse sampling interval $1$, and the range of residual samples was set to $[-1,1]$ with a finer sampling interval $0.1$. For the sparse LFs, only the central view and its adjacent views were used since a large disparity range would lead to high memory consumption. Thus, the range of disparity samples was set to $[-20,20]$ with a coarse sampling interval $1.2$, and the range of residual samples was set to $[-2,2]$ with a finer sampling interval $0.12$. For the OccNet, the maximum channel number is $64$. Therefore, it is very lightweight with only $\SI{0.113}{M}$ parameters.

During training, all the views were cropped to $256\times 256$ randomly. The learning rate was set to $1\times 10^{-3}$ initially, and multiplied by $0.8$ every $50$ epochs. Moreover, we empirically set $\alpha_1=1$ for the SSIM loss, and $\eta=100$, $\alpha_2=0.1$, $\alpha_3=0.05$ for the smoothness losses. The DispNet and OccNet were jointly trained using the loss in Eq.~\ref{eq:full loss} with the Adam optimizer for about $500$ epochs.  

During inference, multiple view combinations were input to the DispNet to obtain multiple disparity maps. For the dense LFs, the input combinations (expressed by the angular coordinates) are $[(u_c-3,v_c),(u_c,v_c),(u_c+3,v_c)]$, $[(u_c-2,v_c),(u_c,v_c),(u_c+2,v_c)]$, $[(u_c,v_c-3),(u_c,v_c),(u_c,v_c+3)]$, $[(u_c,v_c-2),(u_c,v_c),(u_c,v_c+2)]$, and the weighted fusion with $n'=2$ was used to obtain the final disparity map. For the sparse LFs, the input combinations are $[(u_c-1,v_c),(u_c,v_c),(u_c+1,v_c)]$, $[(u_c,v_c-1),(u_c,v_c),(u_c,v_c+1)]$, and the minimum error fusion was used since there are only two estimated disparity maps. The quantile $q$ in Eq.~\ref{eq:M} was set to $0.95$.

\subsection{Comparison}
We compared our method with several state-of-the-art LF depth estimation methods, including both the supervised and unsupervised methods, on a variety of LF datasets. 

\subsubsection{Evaluation on Synthetic Dense LF Images}
First, we list the quantitative results of different methods on several scenes from the HCI dataset in Table~\ref{tab:HCI comparison}, including two supervised methods, EPINet~\cite{Shin2018} and LFAttNet~\cite{Tsai2020}, two non-learning-based methods, OCC~\cite{Wang2016} and FBS~\cite{Lee2017}, and four unsupervised methods, UnCNN~\cite{Peng2018}, UnMonocular~\cite{Zhou2020}, UnPlug~\cite{Iwatsuki2022}, and UnOcc~\cite{Jin2022}. The evaluation metrics are the mean square error (MSE $\times 100$) and bad pixel ratios (BPR)~\cite{Honauer2016} with thresholds $0.07$, $0.03$ and $0.01$ (lower is better for all). The results of the other methods were obtained from the benchmark of the HCI dataset~\cite{Honauer2016} or the original papers~\footnote{The results of UnOcc~\cite{Jin2022} and UnPlug~\cite{Iwatsuki2022} are from their published papers, and they provide only the MSE and BPR with threshold $0.07$.}. It can be seen that our method still has some gaps with the supervised methods but can generally outperform the other unsupervised methods. Some of the visual results (Dino and Sideboard), with the error maps relative to the ground truths, are presented in Fig.~\ref{fig:HCI comparison}, where we can find that our estimated disparity maps have better visual qualities with smaller errors than the other non-learning-based and unsupervised methods. 

Then, we compared the performance of different methods on several scenes from the DLF dataset~\cite{Shi2019}, as recorded in Table~\ref{tab:DLF comparison}. The methods for comparison include EPINet~\cite{Shin2018}, LFAttNet~\cite{Tsai2020}, UnCNN~\cite{Peng2018}, UnMonocular~\cite{Zhou2020}, UnCon~\cite{Lin2022}, and UnOcc~\cite{Jin2022}. We re-built these models and trained them using the same synthetic dense LF datasets since there are no off-the-shelf models for evaluation. From our experiments, we can see that our method obtains better quantitative results than the other unsupervised methods. Compared to the supervised methods, our method still has some gaps in terms of the MSE, but with comparable or lower BPRs. Fig.~\ref{fig:DLF comparison} presents the visual results on the scenes, Toys and Antiques, which reflects that our estimated disparity maps are more visually compelling with proper smoothness and clear object details. 

\subsubsection{Evaluation on Synthetic Sparse LF Images}
To evaluate the performance of different methods on the sparse LF images, we retrained them using the SLF dataset~\cite{Shi2019}. The quantitative results of several scenes in terms of the MSE and BPR with thresholds $0.3$, $0.1$ and $0.05$ are listed in Table~\ref{tab:SLF comparison}. It can be seen that our method achieves comparable quantitative results with the supervised method LFAttNet, and has better results than the others. Both our method and LFAttNet predict the probability distribution of the disparity samples by constructing cost volumes, while the other methods directly output the disparity values from the last convolution layer, which increases the difficulty to learn large disparities for the networks. The visual results on the scenes, Lion and Rooster clock, are shown in Fig.~\ref{fig:SLF comparison}, with a different color map from the dense LFs for distinction. It can be seen that our disparity maps have better visual qualities with smaller errors compared to the other methods. 

\subsubsection{Evaluation on Real-world LF Images}
We further evaluated different methods on the real-world LF images from the datasets~\cite{Raj2016,Kalantari2016,Rerabek2016}. All the methods were trained only on the synthetic LF images. Fig.~\ref{fig:Real comparison} presents several predicted disparity maps of some methods. As the real-world scenes do not have ground truths, we can only compare their visual qualities. It can be seen that our method achieves more smooth predictions while preserving better object details, and therefore demonstrates better robustness and generalization compared to the other methods.  

\subsubsection{Efficiency}

\begin{table}[t]
\caption{Number of parameters and run time of different methods}
\centering
\begin{tabular}{c|cc}
\toprule
Method & Param. (M) & Run time (s) \\
\midrule
EPINet~\cite{Shin2018}      & 2.466 & 2.765 \\
LFAttNet~\cite{Tsai2020}    & 5.540 & 6.479 \\
UnCNN~\cite{Peng2018}       & 1.315 & 5.768 \\
UnMonocular~\cite{Zhou2020} & 1.878 & 1.434 \\
UnOcc\cite{Jin2022}         & 1.891 & 1.840 \\
UnCon~\cite{Lin2022}        & 2.185 & 2.102 \\
UnPlug~\cite{Iwatsuki2022}  & 2.466 & 2.765 \\
Ours                        & 1.802 & 2.688 \\
\bottomrule
\end{tabular}
\label{tab:efficiency}
\end{table}

We list the number of parameters and run time (the average time for outputting the disparity map for one LF image) of different methods, shown in Table~\ref{tab:efficiency}~\footnote{UnPlug~\cite{Iwatsuki2022} employs the architecture of EPINet~\cite{Shin2018}, so their number of parameters and run time are the same.}. All the methods were implemented on a NVIDIA Tesla P100 GPU. It can be seen that our network is lightweight, and achieves a good balance between the accuracy and efficiency.

\subsection{Ablation Studies}

\begin{figure}[t]
\centering
\includegraphics[width=0.48\textwidth]{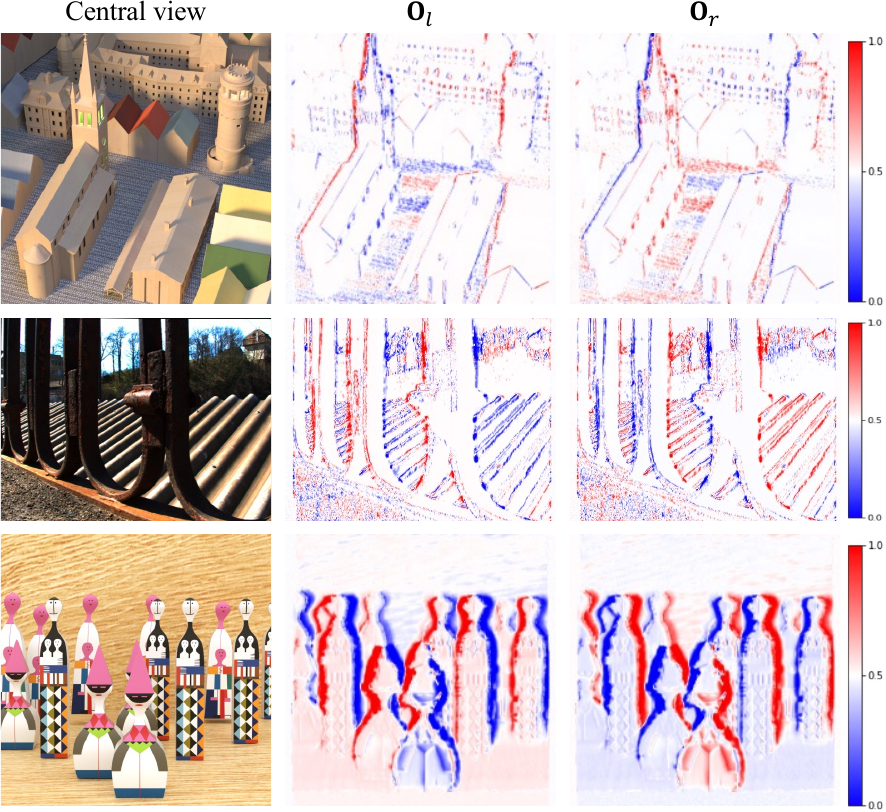} 
\caption{The predicted occlusion maps of a synthetic dense LF, a real-world LF and a synthetic sparse LF.}
\label{fig:occ_maps}
\end{figure}

\begin{figure}[t]
\centering
\includegraphics[width=0.5\textwidth]{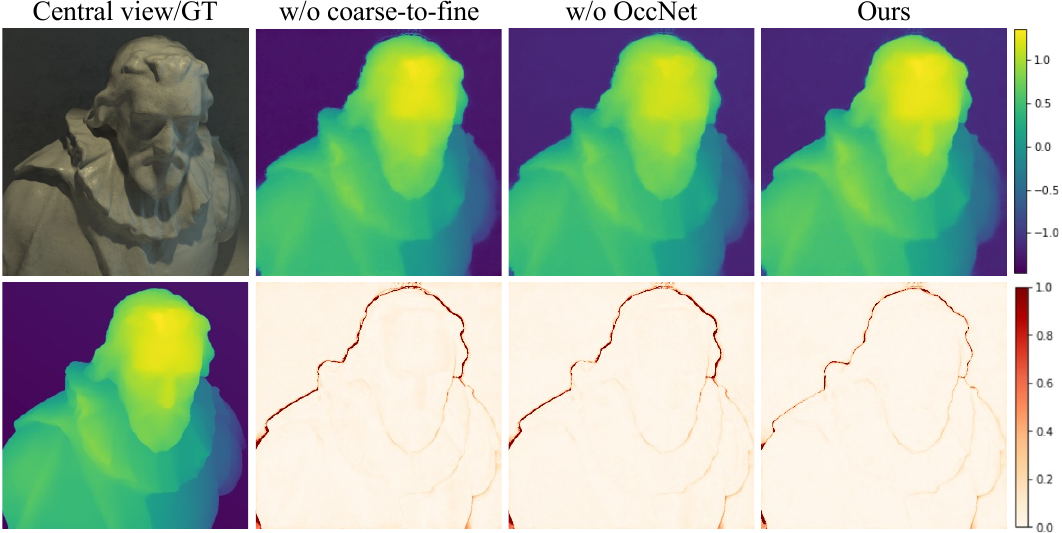} 
\caption{Visual comparison of different network configurations. Zoom in for best view.}
\label{fig:structure_ab}
\end{figure}

\begin{figure}[t]
\centering
\includegraphics[width=0.5\textwidth]{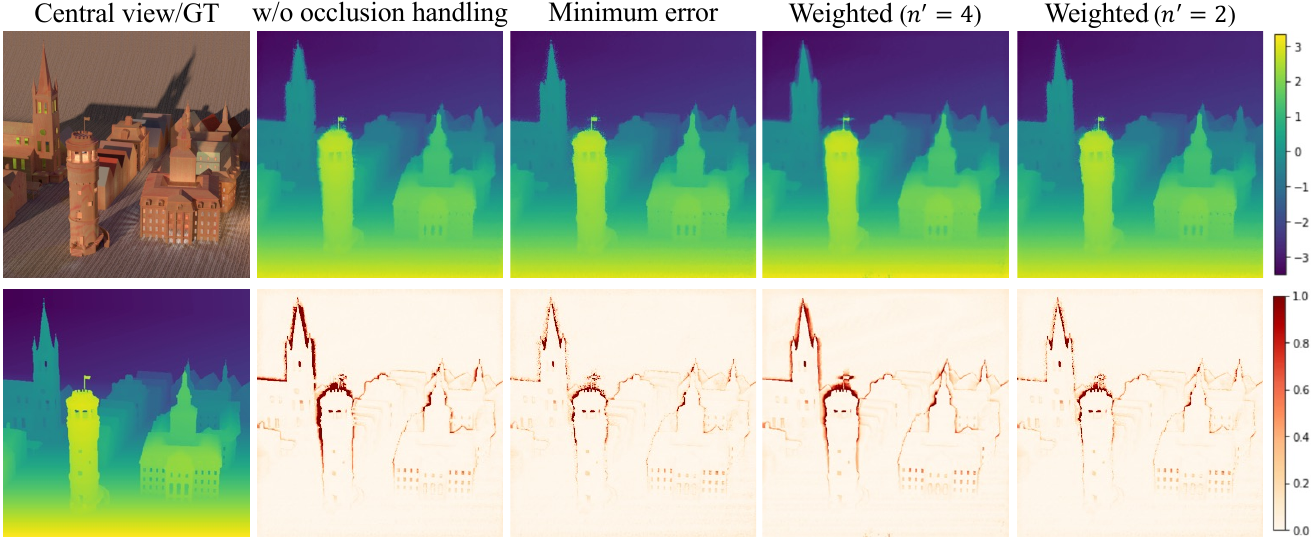} 
\caption{Visual comparison of different disparity fusion strategies. Zoom in for best view.}
\label{fig:fusion_ab}
\end{figure}

\begin{table}[t]
\caption{Ablation study on the framework. Bold: Best.}
\centering
\begin{tabular}{c|c|cccc}
\toprule
\multirow{2}{*}{Configuration} & Param. & MSE$\downarrow$ & BPR$\downarrow$ & BPR$\downarrow$ & BPR$\downarrow$\\[0.5ex]
                          & (M) & ($\times100$) & (0.07) & (0.03) & (0.01)\\
\midrule 
w/o coarse-to-fine & 1.799 & 3.057 & 11.487 & 27.302 & 62.725\\
w/o shared weights & 2.230 & 2.282 & 9.365 & 21.882 & 54.137 \\
w/o OccNet         & 1.802 & 2.701 & 10.806 & 24.466 & 58.244 \\
Default            & 1.802 &\textbf{2.266} &\textbf{9.238} &\textbf{21.683} &\textbf{54.042} \\
\bottomrule
\end{tabular}
\label{tab:structure ablation}
\end{table}

\subsubsection{Design of DispNet}
We first validated the coarse-to-fine structure of DispNet by training an additional model without the branch of residual estimation. We chose several scenes from the HCI and DLF datasets for evaluation. Table~\ref{tab:structure ablation} lists the number of parameters (during inference) and quantitative results of different configurations. It can be seen that the quantitative results have obvious decline if the coarse-to-fine structure is not employed, which suggests that the refinement branch with finer samplings helps to improve the disparity accuracy. Moreover, the refinement branch only introduces extra $\SI{0.003}{M}$ parameters since the feature extractor and the cost filters are all shared by the two branches. The visual comparison in Fig.~\ref{fig:structure_ab} suggests that the DispNet with a coarse-to-fine structure achieves better disparity estimation with smaller errors.

We also trained an additional model without shared weights for the cost filters. From Table~\ref{tab:structure ablation}, we can see that separate cost filters for the two branches lead to more parameters but no improvement on the performance. Therefore, we chose to use shared cost filters in our DispNet. 

\subsubsection{Occlusion Prediction}
To verify the effectiveness of our OccNet for occlusion prediction, we trained an additional model by removing the OccNet and the loss terms $\ell_\mathrm{rec}$ and $\ell_\mathrm{smo}$ in Eq.~\ref{eq:full loss}. Thus, the pixel-wise weight of the photometric loss in Eq.~\ref{eq:wpm loss} was reduced to $0.5$. From Table~\ref{tab:structure ablation}, we observe that the performance is degraded if the OccNet is not employed. The visual comparison in Fig.~\ref{fig:structure_ab} reflects that the disparity map with occlusion prediction has smaller errors.

The predicted occlusion maps for a synthetic dense LF, a real-world LF and a synthetic sparse LF are presented in Fig.~\ref{fig:occ_maps}. The occlusion regions are usually near the object boundaries~\cite{Meng2021}, which leads to larger or smaller values within these regions in the occlusion maps. The red pixels with values approximate to $1$ denote larger contributions for the reconstruction of the central view and also larger weights for the photometric loss. The case is opposite for the blue pixels with values approximate to $0$. Moreover, due to the larger disparity range, the sparse LFs usually have larger occlusion regions, resulting in much thicker red and blue regions near the object boundaries compared to the dense LFs. 

\begin{table}[t]
\caption{Ablation study on the loss terms. Bold: Best.}
\centering
\begin{tabular}{c|cccc}
\toprule
\multirow{2}{*}{Loss terms} & MSE$\downarrow$ & BPR$\downarrow$ & BPR$\downarrow$ & BPR$\downarrow$\\
                          & ($\times100$) & (0.07) & (0.03) & (0.01)\\
\midrule
w/o $\ell_\mathrm{SSIM}$  & 2.393 & 9.535 & 22.027 & 55.250 \\
w/o $\ell_\mathrm{smd}$   & 2.457 & 9.749 & 22.136 & 55.856 \\
w/o $\ell_\mathrm{smo}$   & 2.276 & 9.884 & 22.304 & 54.275 \\
Full loss                 &\textbf{2.266} &\textbf{9.238} &\textbf{21.683} &\textbf{54.042} \\
\bottomrule
\end{tabular}
\label{tab:loss ablation}
\end{table}

\subsubsection{Loss Terms}
We trained additional models by excluding the SSIM loss $\ell_\mathrm{SSIM}$, the smoothness loss for disparity map $\ell_\mathrm{smd}$ and the smoothness loss for occlusion map $\ell_\mathrm{smo}$, since $\ell_\mathrm{wpm}$ and $\ell_\mathrm{rec}$ are indispensable to the training of DispNet and OccNet. Table~\ref{tab:loss ablation} lists the corresponding results, which suggests that the MSE and BPRs slightly increase after removing each of them. Therefore, these loss terms contributes to a better performance.

\begin{table}[t]
\caption{Ablation study on disparity fusion approaches. Bold: Best.}
\centering
\begin{tabular}{c|cccc}
\toprule
\multirow{2}{*}{Fusion strategy} & MSE$\downarrow$ & BPR$\downarrow$ & BPR$\downarrow$ & BPR$\downarrow$\\[0.5ex]
                          & ($\times100$) & (0.07) & (0.03) & (0.01)\\
\midrule
w/o occlusion handling  & 2.837 & 10.371 & 23.441 & 56.778 \\
Minimum error fusion    & 2.411 & 10.210 & 23.743 & 57.842 \\
Weighted fusion $(n'=4)$  & 2.564 & 10.022 & 22.513 & 54.462 \\
Weighted fusion $(n'=2)$  &\textbf{2.266} &\textbf{9.238} &\textbf{21.683} &\textbf{54.042} \\
\bottomrule
\end{tabular}
\label{tab:fusion ablation}
\end{table}

\subsubsection{Disparity Fusion Strategy}\label{sec:fusion}

\begin{figure}[t]
\centering
\includegraphics[width=0.48\textwidth]{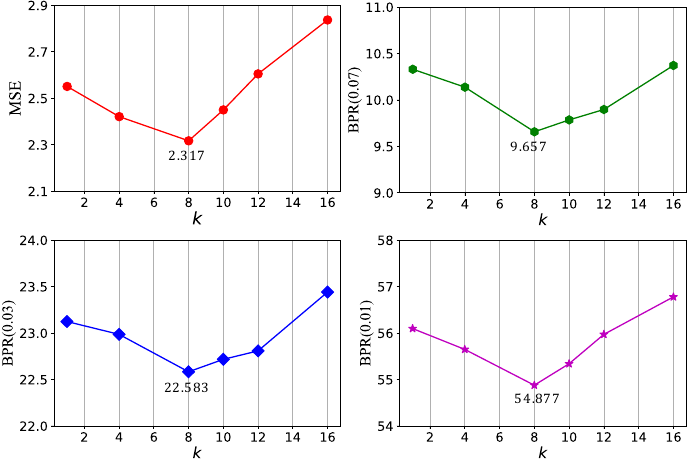} 
\caption{The effect of $k$ on the quantitative results.}
\label{fig:k}
\end{figure}

\begin{figure}[t]
\centering
\includegraphics[width=0.48\textwidth]{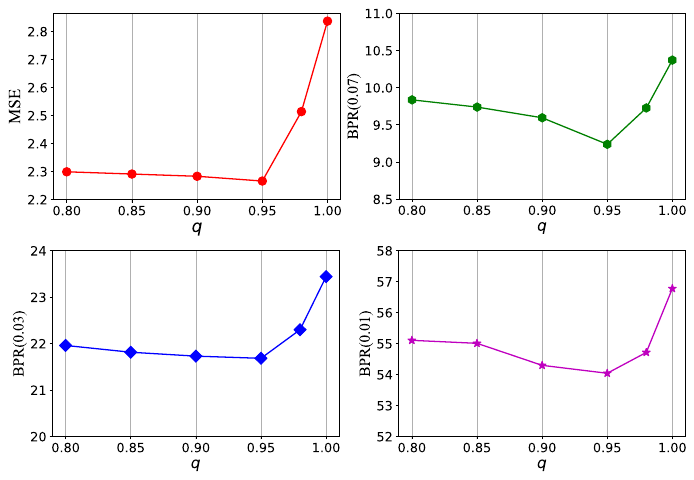} 
\caption{The effect of $q$ on the quantitative results.}
\label{fig:q}
\end{figure}

We first evaluated different disparity fusion strategies, and our default strategy is the weighted fusion by using two disparities ($n'=2$) with the smallest errors at each position. The other fusion approaches include the weighted fusion by using all the disparities $(n'=4)$, the minimum error fusion that selects the disparity with the minimum error at each position. We also list the results of no occlusion handling when deriving the error maps (the estimated error of each position is represented by the mean of all the warping errors) with weighted fusion ($n'=2$). Their quantitative results are given in Table~\ref{tab:fusion ablation} and the visual comparison is shown in Fig.~\ref{fig:fusion_ab}, where we observe that the weighted fusion with $n'=2$ obtains better results than the other strategies, and the performance has obvious decline if occlusion is not handled during fusion.

Another approach to address occlusion is to use the average value of the $k$ smallest warping errors. We experimented with $k=1,4,8,10,12,16$ (totally $16$ auxiliary views) using weighted fusion with $n'=2$. Fig.~\ref{fig:k} shows the quantitative results at different $k$ values. It can be seen that the results are best at $k=8$, but they are still worse than those of using median value as listed in Table~\ref{tab:fusion ablation}. As large warping errors are mainly caused by the occlusion and inaccurate disparity estimation, large $k$ value may lead to incomplete occlusion handling while small $k$ value may lead to inaccurate error estimation at the pixels with inaccurate disparity values. Therefore, using median value is more effective to address these issues.

We then investigated the effect of quantile $q$ for determining the threshold of occlusion. We experimented with $q=0.8,0.85,0.9,0.95,0.98,1$. $q=0.95$ means that $5\%$ pixels with the largest standard deviations use the median value of the warping errors to address the occlusion issue, and $q=1$ means that all the pixels use the mean value of the warping errors, which is equivalent to the strategy without occlusion handling in Table~\ref{tab:fusion ablation}. Fig.~\ref{fig:q} shows the effect of $q$ on the quantitative results, where we can see that $q=0.95$ obtains better results than the other values, and the MSE increases significantly at $q=0.98$ and $q=1$, verifying the effectiveness of occlusion handling.

\section{Conclusion}\label{sec:conclusion}
This paper presents an unsupervised framework for LF depth estimation. We first develop a DispNet that takes as inputs different view combinations to predict multiple disparity maps. It adopts a coarse-to-fine structure with two branches to estimate the coarse and residual disparity maps, respectively, through multi-view feature matching. Photo-consistency is the main supervisory signal for the unsupervised training. However, it does not hold in the occlusion regions. To tackle this issue, we introduce an OccNet to predict the occlusion maps, which are used as the element-wise weights of the photometric loss to alleviate the impact of occlusions on the disparity learning. The OccNet is trained by the reconstruction loss with no ground truth required. The multiple estimated disparity maps after rotating and scaling are fused based on their estimated errors using the auxiliary views with effective occlusion handling to obtain the final disparity map. Our framework achieves superior performance on both the dense and sparse LF images, and also demonstrates better generalization ability on the real-world LF images.


\end{document}